\documentclass{article}

\usepackage[final]{neurips_2019_ml4ad}

\usepackage[utf8]{inputenc} % allow utf-8 input
\usepackage[T1]{fontenc}    % use 8-bit T1 fonts
\usepackage{hyperref}       % hyperlinks
\usepackage{url}            % simple URL typesetting
\usepackage{booktabs}       % professional-quality tables
\usepackage{amsfonts}       % blackboard math symbols
\usepackage{nicefrac}       % compact symbols for 1/2, etc.
\usepackage{microtype}      % microtypography

\usepackage{amsmath}
\usepackage{caption}
\usepackage{subcaption}

\usepackage{graphicx}

\title{Single-step Options for Adversary Driving}

\author{%
  Nazmus Sakib\\
  Department of Computing Science\\
  University of Alberta\\
  \texttt{nazmus@ualberta.ca} \\
  \And
  Hengshuai Yao\\
  Huawei Edmonton R \& D\\
  Huawei Technologies Canada\\
  \texttt{hengshuai.yao@huaweicom}\\
  \And
  Hong Zhang\\
  Department of Computing Science\\
  University of Alberta\\
  \texttt{hzhang@ualberta.ca}\\
  \And
  Shangling Jui\\
  Huawei Kirin Solution\\
  Huawei Technologies Canada\\
  \texttt{jui.shangling@huawei.com}\\
}

\begin{document}

\maketitle

\begin{abstract}
  %Adversary scenarios in driving,  where the other vehicles may make mistakes or have a competing or malicious intent,  have to be studied  not only for our safety but also for addressing the concerns from public in order to push the technology forward.  
%The state-of-art solutions for adversary driving do not exist so far, especially when the vehicles do not communicate their intent. 
In this paper, we use reinforcement learning for safety driving in adversary settings. 
In our work, the knowledge in state-of-art planning methods is reused by single-step options whose action suggestions are compared in parallel with primitive actions. 
We show two advantages by doing so. 
First, training this reinforcement learning agent is easier and faster than training the primitive-action agent. 
Second, our new agent outperforms the primitive-action reinforcement learning agent, human testers  as well as the state-of-art planning methods that our agent queries as skill options. 
\end{abstract}

\section{Introduction}

In this paper, our problem context is autonomous driving. 
The question for us to explore in the long term is, can computers equipped with intelligent algorithms achieve superhuman level safe driving?
The task of autonomous driving is very similar to game playing in the sequential decision making nature. 
Although driving is not a two-player game leading to a final win or loss, accident outcomes can still be treated as loss. 
In driving, an action to take at every time step influences the resulting state which the agent observes next, which is the key feature of many problems where reinforcement learning has been successfully applied. 
However, unlike games, driving poses a unique challenge for reinforcement learning with the stringent safety requirement. 
Although the degree of freedom is relative small for vehicles, the fast moving self-motion, high-dimensional observation space and highly dynamic on-road environments pose a great challenge for artificial intelligence (AI). 
%Another challenge is the disrepancy of simulator and real roads. 
%Attack this problem is challenging because of experimenting with real car can cause accidents on the roads.
%Building simulators and training models via simulation is a popular practice.  

Human drivers drive well in normal traffic. However, human beings are not good at handling accidents because a human driver rarely experiences accidents in one's life regardless of the large amount of accident-free driving time. In this regard, the highly imbalanced positive and negative driving samples poses a great challenge for supervised learning approach for training  self-driving cars.      
We believe that the prospect of autonomous driving is using programs to simulate billions of accidents in various driving scenarios.  With a large scale of accident simulation, reinforcement learning, already proven to be highly competitive in large and complex simulation environments, has the potential to develop the ultimately safest driving softwares for human beings. 

This paper studies the problem of adversary driving where vehicles do not communicate with each other about their intent. 
We call a driving scenario {\em adversary} if the other vehicles in the environment can make mistakes or have a competing or malicious intent. 
Adversary driving are rare events but they can happen on the roads from time to time, posing a great challenge to existing state-of-art autonomous driving softwares. 
Adversary driving has to be studied, not only for our safety, but also to address the safety concerns from the public in order to push the technology forward. 
However, most of the state-of-art planning algorithms for autonomous driving do not consider adversary driving, usually assuming all the agents in the environment are cooperative. For example, the other vehicles are assumed to be ``self-preserving'', which actively avoid collisions with other agents whenever possible, e.g., see \citep{pierson2018navigating}. 
Optimal Reciprocal Collision Avoidance (ORCA) \citep{orca}  is a popular navigation framework in crowd simulation and multi-agent system for avoiding collision with other moving agents and obstacles. 
The traffic that ORCA generates is cooperative, by planning on each vehicle's velocity to avoid collision with others. 
Recently, \citep{Abeysirigoonawardena2018GeneratingAD} pointed that there is practical demand to simulate adversary driving scenarios in order to test the safety of autonomous vehicles. 
 
Our work is the first attempt to solve non-communicating adversary driving. 
We use a reinforcement learning approach. 
Driving has a clear temporal nature, {\em the current action has an effect on choosing the actions in the future}. 
Reasoning which action to apply by considering its long-term effects is usually called a temporal credit assignment, 
which is usually modeled as a reinforcement learning problem. 
In most recent reinforcement learning applications, there is a deep neural networks that maps an input state to an optimal policy over primitive actions. However, learning a policy over primitive actions is very difficult and inefficient. For example, hundreds of millions of frames of interacting with the environment are required in order to learn a good policy even for a simple 2D game in Atari 2600.  
In a simulated driving environment, deep reinforcement learning was found to be much inferior to state-of-art planning and supervised learning, in both the performance and the amount of training time \citep{dosovitskiy2017carla}. 

 \begin{figure}
  \centering
  \includegraphics[width=5cm,angle=90]{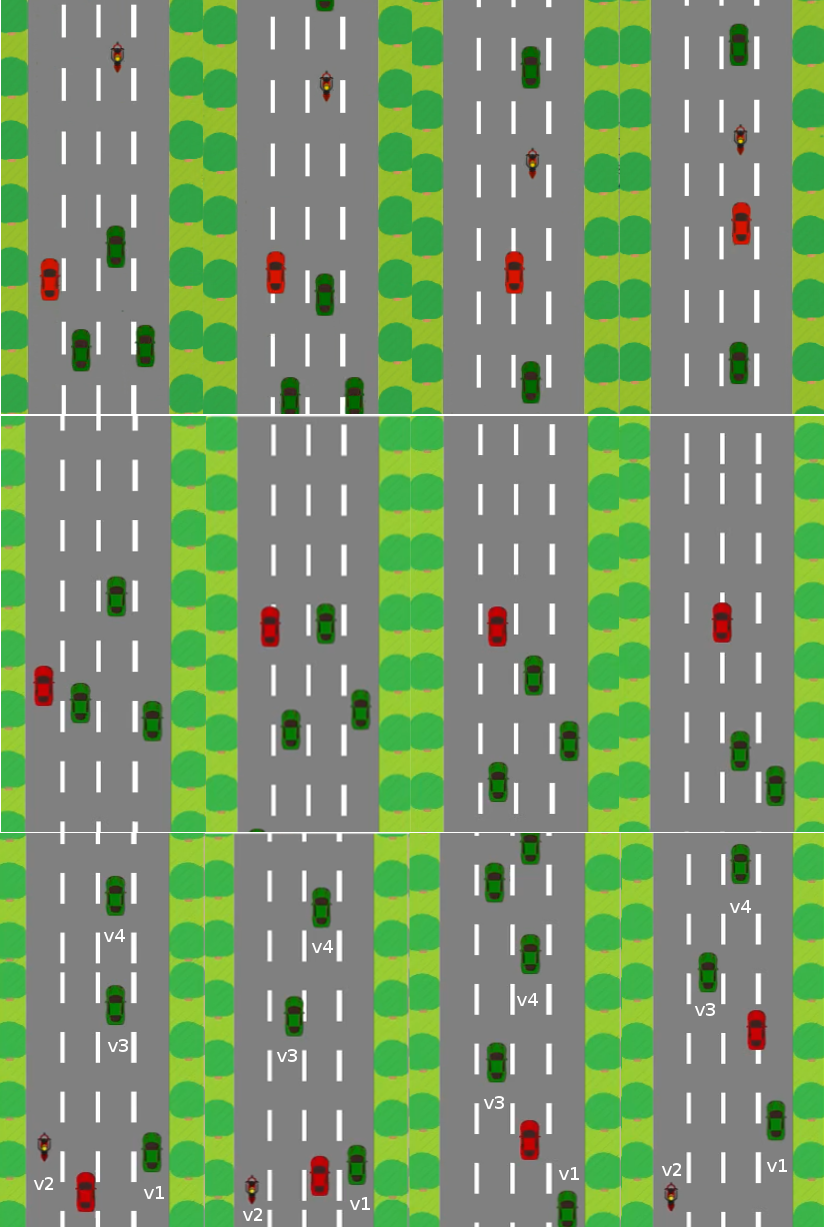}
  \caption{Successful moments of driving with our method: merging (column 1), passing (column 2) and finding gaps (column 3). }
  \label{fig:success}
\end{figure}

%\begin{figure}[t]
%      \centering
%        \includegraphics[angle=90,width=0.47\textwidth]{success.png}
%        \caption{Successful moments of driving with our method: merging (row 1), passing (row 2) and finding gaps (row 3). }
%        \label{fig:success}
% \end{figure}

On the other hand, autonomous driving field has already practised a rich set of classical planning methods. 
It is worth pointing out that the problem of state-of-art planning is not that their intended performance is bad. In fact,  both research and industrial applications have shown that classical planning works great in the scenarios they are developed for. 
The problem of state-of-art planning methods is the existence of logic bugs and corner cases  that were not considered at the time of developing. 
As a result, updating, testing and maintaining these software modules is the true darkness of this method: huge intensive human engineering labor is involved. 
However, the knowledge already learned in state-of-art planning methods should be inherited and reused. 
Implemented state-of-art planning softwares in the autonomous driving industry have logic bugs but they perform well in the general scenarios that they are developed for, usually accounting for perhaps a large percentage of everyday driving.  
How do we reuse these state-or-art planning softwares and solving their logic holes automatically without human engineering?

To summarize, in autonomous driving, the field has implemented a rich set of state-of-art planning methods but they have logic holes that originate from either oversight or mistakes in the process of software engineering. 
Deep reinforcement learning that learns a policy over primitive actions is slow to train but they can explore the action space for the best action though large numbers of simulations. 
The framework of options is used to describe planning with abstract decisions that usually last for more than one time steps. 
An option is usually defined as a policy, an initiation set where the policy can be initiated, and a termination condition upon satisfying to stop executing the policy \citep{options99}. In practice,  the options framework is usually in the form of a hierarchy with a meta-policy at the top, a number of skill models in the middle level(s) which calls upon certain primitive actions in the bottom level. 
Our work in this paper is a novel architecture of options, which features in a flat architecture of single-step options. 
In particular, to take advantage of both state-of-art methods and (flat) end-to-end RL, we define a set of actionable options that can be called every time step by the learning agent. 
In addition to the primitive actions which are are natural single-step options, 
we also enable the agent to call state-of-art planning methods for action suggestions every time step. 
In this way, state-of-art planning methods are treated as single-step options and are reused as skills in the framework of options. 

They can be called with an input state and gives an action suggestion. 
Our reinforcement learning agent will be able to select over the action suggestions by state-of-art planning methods as well as the primitive actions. 
Our method is able to call classical planning methods to apply the skills in normal conditions for which they are developed, 
but is also able to pick the best primitive action to avoid collision in scenarios where state-of-art planning fails to ensure safety. 
In this way, we do not have to re-learn for the majority of scenarios in driving where classical planning methods already can deal with, saving lots of time for training the deep networks, and focus on the rare but most challenging scenarios where they are not designed for. 
The advantage of our method is that we do not have to manually detect whether state-of-art planning fails or not, 
instead, failures of their actions are propagated by reinforcement learning to earlier time steps and remembered through neural networks in training to avoid selecting state-of-art planning on the similar failure cases in the future.

%Our work opens the door to a novel architecture solution for autonomous driving: building a decision hierarchy of skills using state-of-art planning or learning-based methods, and calling them as augmented actions by reinforcement learning. 
To provide contextual background, we discuss state-of-art planning, learning-based control, reinforcement learning, the end-to-end and the hierarchical decision making structure in the remainder of this section. 

\subsection{Classical Planning and Learning-based Control}
We categorize control methods into two classes. 
The first is classical planning, where a control policy is derived from a system model. 
Methods including proportional integral derivative (PID) controllers, model-predictive controller (MPC) and rapidly-exploring random tree (RRT) are examples of classical planning.
The second class is learning-based control, usually used in pair with a deep neural networks. 
Learning-based control is {\em sample based} or {\em data driven}. 
Both supervised learning and reinforcement learning are learning-based methods. 
Supervised learning systems aim to replicate the behavior of human experts. 
However, expert data is often expensive, and may also impose a ceiling on the performance of the systems. 
By contrast, reinforcement learning is trained from explorative experience and is able to outperform human capabilities. 
Note that though, supervised learning can be useful in training certain skills of driving. 
For example, a supervised learning procedure is applied to imitate speed control by human drivers via regulating the parameters in a linear relationship between speed and shock to the vehicle on off-road terrains \citep{stavens2007online}.  

The advantage of state-of-art planning is that algorithms are easy to program, and have good performances though often with significant efforts in parameter tuning. 
For example, ORCA \citep{orca} has many parameters and difficult to tune. A POMDP based planner was proposed in \citep{Intention2015} but is computationally expensive, which runs at 3Hz (our method works at 60Hz). Their environment consists of slowly moving agents, which is not suitable for high speed vehicles. \citet{SIPP2011} used waiting time to reduce the search space of collisions in future time stamps. However, they do not consider actuation control (acceleration/deceleration). However, the method of using waiting time is not applicable in highway. It has also a prediction step for tracking other agents which an extra layer of complexity. The planner by \citet{provablysafemotion} also relies on explicitly predicting the motion of other agents. It uses kinematic model of the non-ego agents and generates the occupancy in future time steps for a safe trajectory. The difference of our method from these two methods is that ours does not explicitly predict other agents’ future state. In this work, we learn a neural networks to implement a policy that maps state directly to optimal actions.

Implementing state-of-art planning methods usually requires significant domain knowledge, and they are often sensitive to the uncertainties in the real-world environment \citep{long2017deep}. 
Learning-based control, on the other hand, enables mobile robots to continuously improve their proficiency and adapt to the statistics of real-world environments with the data collected from human experts or simulated interactions. 

\subsection{Classical Planning and Reinforcement Learning}
Classical planning methods have already been widely adopted in autonomous driving. 
Recent interests in using reinforcement learning also arise in this new application field. 
We comment that this is not incidental. 
Specifically, there are a few common fundamental principles in the core ideas of classical planning and reinforcement learning. 

First, temporal relationship between the actions selected at successive time steps is considered in both fields. 
Optimizing the cost over future time steps is the key idea commonly shared between classical planning and reinforcement learning algorithms. 
For example, in MPC, there is a cost function defined over a time horizon for the next few actions. 
The cost function is one special case of the (negative) reward function in reinforcement learning. 
MPC relies a system model and an optimization procedure to plan the next few optimal actions.
The collision avoidance algorithm using risk level sets maps the cost of congestion to a weighed graph along a planning horizon, and apply Djikstra's Algorithm to find the fastest route through traffic \citep{pierson2018navigating}. 
Many collision avoidance planning algorithms evaluate the safety of the future trajectories of the vehicle by predicting the future motion of all traffic participants, e.g., see  \citep{lawitzky2013interactive}.
However, MPC, Djikstra's Algorithm and collision avoidance planning are not sampled based, while reinforcement learning algorithms are sample-based. 

Second, both fields tend to rely on decision hierarchies for handling complex decision making. 
Arranging the software in terms of high-level planning, including route planning and behavior planning, and low-level control, including motion planning and closed-loop feedback control became a standard for autonomous driving field \citep{urmson2007tartan,montemerlo2008junior,mobileye-longshort-16}. 
In reinforcement learning, low-level options and a high-level policy over options are separately learned \citep{option-critic}. 
In robotics, locomotion skills are learned at a fast time scale while a meta policy of selecting skills is learned at a slow time scale \citep{deeploco}. 

Third, sampling-based tree search methods exist in both fields. 
For example, RRT is a motion planning algorithm for finding a safe trajectory by unrolling a simulation of the underlying system \citep{motion-plan-emilio-08}. In reinforcement learning, Monte-Carlo Tree Search (MCTS) runs multiple simulation paths from a node to evaluate the goodness of the node until the end of each game.

%\subsection{Autonomous Driving and Reinforcement Learning}
%The prospect of developing a superhuman level driving agent using reinforcement learning is intriguing. 
%Training reinforcement learning-equipped vehicles on the road is not practical, and so far the efforts have been mainly in simulators. 
%A truthful and easily configurable simulator is crucial to develop reinforcement learning-based agents. 
%The literature has seen many recent efforts on this. 
%For example, 
%CARLA is an open-source driving simulator that is specifically designed to develop for training learning-based agents \citepp{dosovitskiy2017carla}, with a benchmark comparison of both supervised learning and reinforcement learning agents.
%DeepTraffic implements a reinforcement learning agent for lane control in  high-way driving, and hosts a competition for fine tuning the parameters for their implemented algorithm \citepp{deeptraffic}. 
%In this paper, we also developed a high-way lane control simulator that can be easily customized. 
%
%A small wheel-vehicle  trained with reinforcement learning navigate with collision avoidance in a pedestrian-rich environment \citepp{chen2017socially}. 
%Long-term driving strategies are generated by a hierarchical temporal abstraction graph \citepp{shalev2016safe}. 
%A robot RC car was trained to navigate in an complex indoor environment using an efficient method that predicts multiple steps on a computation graph \citepp{kahn2018self}.

\subsection{End-to-End and Hierarchical Decision Making}
The end-to-end approach is the state-of-art architecture for learning-based agents, with remarkable success in hard AI games due to reinforcement learning \citep{dqn,alphaGo0-2017} and well practised with supervised learning for high-way steering control in autonomous driving \citep{dave,alvinn,nvidia-2016}.
Such end-to-end systems usually use a deep neural networks that takes in a raw, high-dimensional state observation as input and predicts the best primitive action at the moment. The end-to-end approach is flat, containing only {\em a single layer of decision hierarchy}.    
On the other hand, there is also evidence that most autonomous driving architectures follow a hierarchical design in the decision making module \citep{montemerlo2008junior,urmson2007tartan}. 

Our insight is that hierarchical decision making structure is more practical for autonomous driving. 
Imagine a safety driver sitting at the back of the wheel in a self-driving car. 
Monitoring an end-to-end steering system and reading the numerical steering values in real time is not practical for a fast intervention response. 
However, a hierarchically designed steering system can tell the safety driver a keeping-lane behavior is going to occur in the next two seconds. It is easier to monitor such behaviors in real time, and is able to interrupt timely in emergent situations. 
Not only for safety drivers, system designers need to understand autonomous behaviors in order to improve programs.
Future passengers will also be more comfortable if they can understand the real-time behavior of the vehicle they are sitting in. 
%Decomposing decision making in driving seems a most natural choice for human drivers too. 

However, the current hierarchical design of the decision making module in driving softwares is highly rule and heuristic based, for example, the use of finite-state-machines for behavior management \citep{montemerlo2008junior}, and heuristic search algorithms for obstacle handling \citep{montemerlo2008junior}. Remarkably similarly, these algorithms have also dominated in the early development of many AI fields, yet they were finally outperformed in Chess \citep{lai2015giraffe,silver2017mastering}, Checker \citep{samuel1959some,chellapilla2001evolving,schaeffer2007checkers}, and Go \citep{silver-thesis}, by  
reinforcement learning agents which use value function to evaluate states and training the value function using temporal difference methods aiming to achieve the largest future rewards \citep{sutton-thesis}. 
The paradigm of playing games against themselves and with zero human knowledge in the form of rules or heuristics has helped reinforcement learning agents achieving superhuman-level performance \citep{tesauro1995temporal,alphaGo0-2017}. 

The remainder of this paper is organized as follows. 
Section \ref{sec:method} contains the details about our method. 
In Section \ref{sec:experiment}, we conduct experiments on lane changing in an adversary setting where the other vehicles may not give the way. 
%Section \ref{sec:future}. 
Section \ref{sec:conclusion}  discusses future work and concludes the paper.

\section{Our Method}\label{sec:method}

%  \subsection{Reinforcement Learning}
        We consider a Markov Decision Process (MDP) of a state space $\mathcal{S}$, an action space $\mathcal{A}$, a reward ``function'' $R: \mathcal{S} \times \mathcal{A} \rightarrow \mathbb{R}$, a transition kernel $p: \mathcal{S} \times \mathcal{A} \times \mathcal{S} \rightarrow [0, 1]$, and a discount ratio $\gamma \in [0, 1)$. In this paper we treat the reward ``function'' $R$ as a random variable to emphasize its stochasticity. Bandit setting is a special case of the general RL setting, where we usually only have one state.
        
        We use $\pi: \mathcal{S} \times \mathcal{A} \rightarrow [0, 1]$ to denote a stochastic policy. We use $Z^\pi(s, a)$ to denote the random variable of the sum of the discounted rewards in the future, following the policy $\pi$ and starting from the state $s$ and the action $a$. We have $Z^\pi(s, a) \doteq \sum_{t=0}^\infty \gamma^t R(S_t, A_t)$, where $S_0 = s, A_0 = a$ and $S_{t+1} \sim p(\cdot | S_t, A_t), A_t \sim \pi(\cdot| S_t)$. The expectation of the random variable $Z^\pi(s, a)$ is $$Q^\pi(s, a) \doteq \mathbb{E}_{\pi, p, R}[Z^\pi(s, a)]$$ which is usually called the state-action value function. 
        In general RL setting, we are usually interested in finding an optimal policy $\pi^*$, such that $Q^{\pi^*}(s, a) \geq Q^\pi(s, a)$ holds for any $(\pi, s, a)$. All the possible optimal policies share the same optimal state-action value function $Q^*$, which is the unique fixed point of the Bellman optimality operator (\citep{bellman2013dynamic}),
        
        \begin{align*}
        Q(s, a) = \mathcal{T}Q(s, a) \doteq \mathbb{E}[R(s, a)] + \gamma \mathbb{E}_{s^\prime \sim p}[\max_{a^\prime} Q(s^\prime, a^\prime)]
        \end{align*}

{\bfseries Q-learning and DQN}. 
        Based on the Bellman optimality operator, \citep{watkins1992q} proposed Q-learning to learn the optimal state-action value function $Q^*$ for control. At each time step, we update $Q(s,a)$ as 
        \begin{align*} 
        Q(s, a) \leftarrow Q(s, a) + \alpha (r + \gamma \max_{a^\prime}Q(s^\prime, a^\prime) - Q(s, a))
        \end{align*}
        where $\alpha$ is a step size and $(s, a, r, s^\prime)$ is a transition. There have been many work extending Q-learning to linear function approximation (\cite{sutton2018reinforcement,szepesvari2010algorithms}). \citep{mnih2015human} combined Q-learning with deep neural network function approximators, resulting the Deep-Q-Network (DQN). Assume the $Q$ function is parameterized by a network $\theta$, at each time step, DQN performs a stochastic gradient descent to update $\theta$ minimizing the loss
        \begin{align*}
        (r_{t+1} + \gamma \max_a Q_{\theta^-}(s_{t+1}, a) - Q_\theta(s_t, a_t)) ^ 2
        \end{align*}
        where $\theta^-$ is target network (\citep{mnih2015human}), which is a copy of $\theta$ and is synchronized with $\theta$ periodically, and $(s_t, a_t, r_{t+1}, s_{t+1})$ is a transition sampled from a experience replay buffer (\citep{mnih2015human}), which is a first-in-first-out queue storing previously experienced transitions.

{\bfseries Augmented action space with classical planning methods}. 
The State-of-art implementation of reinforcement learning  uses an action space over the primitive actions, and a neural networks that maps an input state to a policy over primitive actions. 
To take advantage of classical planning methods, we treat them as action functions that can be queried with a state input and gives an action suggestion. Our method is an implementation of DQN with augmented action space from both primitive actions and action query functions by classical planning methods. 

\section{Experiment}\label{sec:experiment}
Our task is to control an ego vehicle in a lane changing task that moves itself to the rightmost lane without collision. 
This scenario happens frequently when we drive close to freeway exits in everyday life.

\subsection{The Adversary Lane-change Simulator}
The driving simulator consists of 4 lanes in a 2D space. Each lane is subdivided into 3 corridors.
There were 19 vehicles in total within a 200 meter range. 
All the vehicles do not communicate with each other. 
Seven other vehicles can change lane randomly with probability $0.01$ at each time step. When they change lane, there is no safety function applied, which poses a great challenge to control the ego vehicle safely. 
Faster vehicles than the ego vehicle disappear from the top of the window and then reappear at the bottom at random lanes with a random speed ranging from 20 km/h to 80 km/h. In this way we ensure a diverse traffic congestion.
Vehicle types include car and motorcycle. 
A car occupies three corridors and a motorcycle occupies one corridor. 
We map the pixels of simulator into meters. A car is represented as a $(width=2m, height=4m)$ rectangle and a motorcycle is a $(0.6m, 1.5m)$ rectangle.

{\em State Representation}.
We used occupancy grid as state representation \citep{deeptraffic}. 
The grid columns correspond to the current lane plus two lanes on each of the left and right sides of the car. 
At each time step (16 ms), the simulator returns the observations of the positions, speeds, distances of the other vehicles in ego-centric view. It also returns collision and safety breaking events. 
(We set  the safety distance threshold to two meters from the front and back of the ego vehicle.)
 
Along the y-direction, we take $50$ meters in the front and $50$ meters in the back of the ego car and discretize the zone with one meter per cell resulting in a grid of shape $(5, 100)$. The value of a cell is the speed of the vehicle in the cell; if no occupying vehicle in the cell, the cell value is zero. 
%For the empty cells and the cells containing the ego vehicle, we subtract their values with the speed limit ($100.0$). 
We normalize the occupancy grid matrix by diving with the speed limit. 

{\em Reward}. 
Whenever the ego agent reaches the rightmost lane, a positive reward of $10.0$ is observed. 
For collisions, a $ -10.0$ negative reward is given. 
For each safety distance breaking event, a negative reward of $ -1.0$ is observed. 
If the agent fails in reaching the rightmost lane within $8,000$ simulation steps,  a negative reward of $-10.0$ is given. 
A constant reward $-0.001$ is given at the other time steps to encourage reaching the goal quickly.

Reinforcement learning agents need to interact with the simulator continuously through episodes. 
For each interaction episode, we initialize the ego car at the leftmost lane. 
An episode is terminated if reaching the rightmost lane successfully or fails with a collision or safety breaking.

{\em Classical Planning Methods}. 
To the best of our knowledge, there is no state-of-art classical planning methods that work for this non-communicating adversary driving scenario.  
We implemented three planning methods by assuming all the other vehicles do not change lane. 

Method P1: If there are sufficient gaps in the front of the ego vehicle in the current lane and both the front and back in the right lane, switch right; otherwise, follow the front vehicle in the current lane with a PID controller for a target speed. 
If there is no vehicle in the front and the right lane is occupied, a target speed of the speed limit is applied. 

Method P2: 
This method is more complex than Method P1. It mimics advanced human driving by checking both the gaps in the right lane and the speed of the closest cars in the right lane, to ensure that none of the vehicles will run into the breaking distance of the ego vehicle.
%We used the mapping table of ``speeds and stopping distances" by the State of Virginia. 
%\footnote{
%\url{https://law.lis.virginia.gov/vacode/46.2-880/}
%} 

Method P3 is an implementation of the risk level sets \citep{pierson2018navigating}. 
For the correctness check of our implementation, 
we tested it in a simplified scenario where all the other vehicles do not change lane. 
We noted that our implementation was able to ensure collision free driving as claimed in their paper. 

Note that Methods P2 and P3 are just for reader's information about how competitive they can be relative to Method P1 in the adversary setting. They are not used as a serious comparison to our method because they are developed for non-adversary settings. 
In our method, Method P1 is used as an action function to augment the action space. 
Without loss of generality, our method can also work with other classical planning methods added into the action space.  

\begin{table}
\caption{{\it The adversary lane changing task: Performance of our method, end-to-end reinforcement learning, human and three planning methods. ``Ours-P'' is the our method with P being the additional single-step option besides primitive actions. } }
\label{table:testing}
\begin{center}
\begin{small}
\begin{tabular}{|c|c|c|c|c||c|c|c|}
\hline
                     & Ours-P1 & Ours-P3         & primitive agent             & human        & P1             & P2                & P3  \\
\hline
collision     & $2.1\%$  &  $2.4\%$ & $6.0\%$   &  $16.0\%$  &  $14.2\%$   & $11.6\%$   & $9.9\%$ \\
\hline 
success     & $85.0\%$ & 81.3\% & $70.1\%$ & $79.2\%$  & $69.4\%$ & $69.6\%$   &  $71.7\%$ \\ 
\hline
avr. speed   & $54.7$   & 51.5& $57.6$     &  $48.0$       & $55.2$      &  $54.1$        & $58.0$ \\
\hline
\end{tabular}
\end{small}
\end{center}
\end{table}

\subsection{Algorithm Setup}
{\em Action Space}. 
For the ``primitive'' agent, the actions are ``accelerate'', ``no action'', ``decelerate'' and ``switch right''. 
The ``accelerate'' action applies a constant acceleration of $3m/s^2$. 
The ``decelerate'' action applies a deceleration of $4 m/s^2$. 
The ``no action'' applies no action and the momentum of the car is kept. 
The ``switch right'' action will start changing to the right lane lane with a fixed linear speed. 
It requires a few simulator steps in order to reach the right lane.  
For our method, the action space is augmented with Method P1. 

The primitive DQN agent's neural network: 
The input layer has the same size as the state occupancy grid. 
There are three hidden layers, each of them having 128 neurons with the ``tanh'' activation function. 
The last layer has $4$ (the number of actions) outputs, which is the Q values for the four actions given the state. 
The learning rate is $10^{-4}$, the buffer size for experience replay is $10^6$, the discount factor is $0.99$, and the target network update frequency is $100$. 
An epsilon-greedy strategy for exploration was used for action selection.
With probability $\epsilon$, a random action is selected. 
With probability $1-\epsilon$, the greedy action, $a^* = \arg\max_{a\in \mathcal{A}} Q(s, a)$ is selected at a given state $s$. 
In each episode, the value of $\epsilon$ starts from $0.1$ and diminishes linearly to a constant, $0.02$. 

Our method is also implemented with a DQN agent, which has the same neural networks architecture as the primitive agent, except that the output layer has $5$ outputs, which include the Q values for the four same actions as the primitive agent plus the Q value estimate for Method P1. 
The learning rate, buffer size, and discount factor, target network update frequency and exploration is completely the same as the primitive agent. 

\subsection{Results}
\begin{figure}
     \centering
     \begin{subfigure}[b]{0.4\textwidth}
         \centering
         \includegraphics[width=\textwidth]{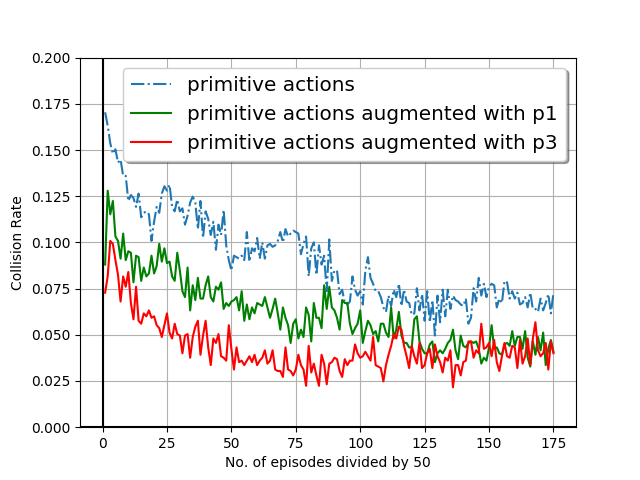}
         %\caption{$y=x$}
         %\label{fig:y equals x}
     \end{subfigure}
     \begin{subfigure}[b]{0.4\textwidth}
         \centering
         \includegraphics[width=\textwidth]{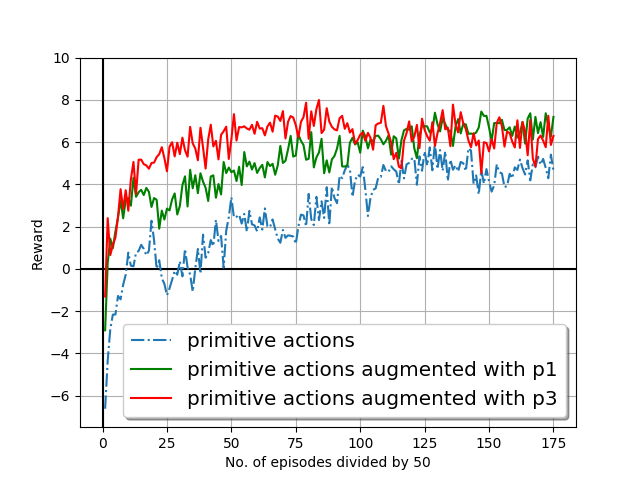}
         %\caption{$y=3sinx$}
         %\label{fig:three sin x}
     \end{subfigure}
        \caption{Learning curves: our method (an RL agent with primitive actions and super skill actions) vs. flat (an RL agent with only the primitive actions). Left: collision rate. Right: reward.}
        \label{fig:training}
\end{figure}

Figure \ref{fig:training} (left) shows the learning curves. 
For every 50 episodes, we computed the collision rate.  
Thus the x-axis is the number of training episodes divided by $50$. 
The y-axis shows the collision rate in the past 50 episodes.
The curves show that our method learns much faster than the primitive agent. 
With the augmented planning method (Method P1) providing action suggestion, 
we effectively reduce the amount of the time and samples in order to learn a good collision avoidance policy.  
Figure \ref{fig:training} (right) shows that our method also learns larger rewards in the same amount of training time.

% \begin{figure}[t]
%     \centering
%     \begin{subfigure}[b]{0.43\linewidth} %%or \columnwidth
%         \centering
%         \includegraphics[width=0.4\linewidth]{rsc/collision_CORL}
%     \end{subfigure}
%     \begin{subfigure}[b]{0.43\linewidth} %%or \columnwidth
%         \centering
%         \includegraphics[width=0.4\linewidth]{rsc/reward_CORL}
%     \end{subfigure}
%     \caption{Learning curves: our method (an RL agent with primitive actions and super skill actions) vs. flat (an RL agent with only the primitive actions). Left: collision rate. Right: reward. }
%  \label{fig:training}
% \end{figure}

We also tested the final performance after training finishes in $10,000$ episodes for both the primitive agent and our agent.  
In addition, we also implemented a gaming system using Logitech G29 consisting driving wheels, acceleration and deceleration paddles, to collect human performance data. % as shown in Figure \ref{fig:system}. 
Three human testers were recruited. 
Each tester was trained for 30 minutes. 
Their best performance over 30 trials was recorded. 
In each trial, 25 episodes were attempted. 
Finally, their performances were averaged to get the human performance index. 

Table \ref{table:testing} shows the performance of our method compared to the primitive agent and human. 
Our method performs better than both the primitive agent and human, achieving a low collision rate of $2.1\%$. 
This low rate was achieved with a similar average speed to primitive agent and human. 
In terms of the rate of successfully reaching the rightmost lane within the limited time, our algorithm achieves $85.0\%$, which is much higher than primitive agent ($70.1\%$) and human ($79.2\%$). 
It seems human testers tend to drive at slow speeds to reach a good success rate. 
Because collision is unavoidable in this adversary setting, the performance of our method is very impressive. 
Note that in the end of training shown in Figure \ref{fig:training}, the collision rate of our method was around $4\%$ instead of being closing to our testing performance, $2.1\%$. 
This is due to that in the end of training, there is still a random action selection with probability of $0.02$ used in epsilon-greedy exploration. 

The table also shows the collision rate of Method P1 is $14.2\%$ on this adversary setting. 
This poor performance is understandable because Method P1 was developed in a much simpler, non-adversary setting. 
The interesting finding here is that by calling Method P1 in our method as augmented action, 
we learn to avoid collision faster as well as improve the collision rate of Method P1 significantly by using reinforcement learning for action exploration.  
Thus our method achieves the goal of reusing classical planning as skills to speed up learning. 
The other planning methods P2 and P3, although perform better than P1, still cannot solve the adversary task with a satisfactory performance.

Figure \ref{fig:success} shows the successful moments of driving with our agent. 
The first column shows a sequence of actions applied by our agent that successfully merge in between two vehicles on the right. 
Specifically, the first moment accelerates; the second moment cuts in front of the vehicle on the right;
and the third and forth moments merge in between two other vehicles on the right. 
The second column shows our agent speeds up and successfully passes other vehicles on the right.
The third column, helped with annotations of the surrounding vehicles. 
In the first moment, our vehicle is looking for a gap. 
The second moment,  {\em v3} switches left, creating a gap and the ego car switches right into the gap.
In the following moments, the ego car keeps switching right because there are gaps on the right.

%\begin{figure}[t]
%      \centering
%        \includegraphics[width=0.15\textwidth,  angle =90]{system_small.png}
%        \caption{The system (rotated 90 degrees) used for collecting driving performance data from human testers: A logitech driving wheel, acceleration and braking paddles, and a chair.}
%        \label{fig:system}
% \end{figure}

\subsection{Knowledge Learned for Driving}
\begin{figure}
     \centering
     \begin{subfigure}[b]{0.3\textwidth}
         \centering
         \includegraphics[width=\textwidth]{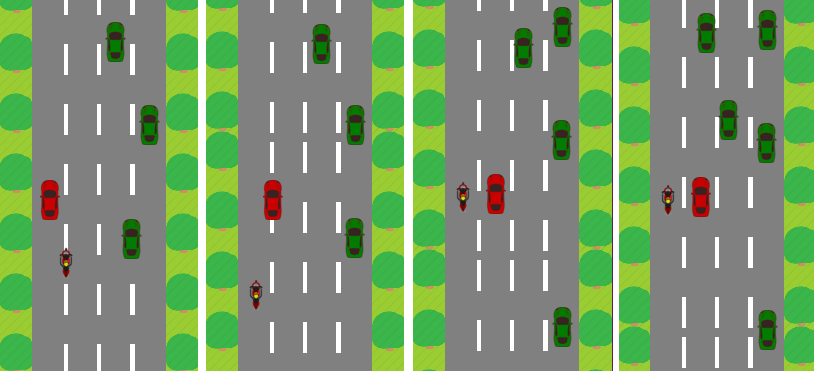}
         %\caption{$y=x$}
         %\label{fig:y equals x}
     \end{subfigure}
     \begin{subfigure}[b]{0.2\textwidth}
         \centering
         \includegraphics[width=\textwidth]{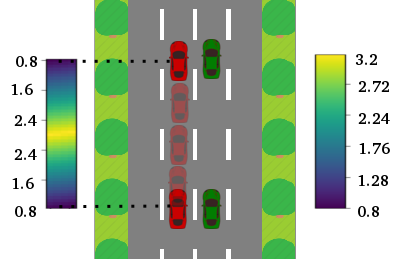}
         %\caption{$y=3sinx$}
         %\label{fig:three sin x}
     \end{subfigure}
        \caption{
        {\small 
            Left : Sampled moments: Q values for the actions. 
                In the order of ``accelerate'', ``no action'', ``deceleration'', ``switching right'':
        	the first moment (accelerating), the action values are,  $[0.851, 0.841, 0.829, 0.844]$;
        	the second moment (decelerating), the action values are, $[1.030, 1.042, 1.043, 1.036]$; 
        	the third moment (accelerating), the action values are, $[1.421, 1.416, 1.406, 1.418]$
        	and the fourth moment (decelerating), the action values are, $[1.316, 1.324, 1.334, 1.319]$. Right : The left color plot shows the values of switching right within the time window:
        the middle moments have the largest values for switching right; while at the two ends, the values are small, indicating the switching right is not favorable because collision will occur. 
        The right color bar is the color legend. 
        The middle shows the trace of the car in the time window that corresponds to the left color plot (dotted line). 
        It shows that the best moment to switch right is near the middle line of the two vehicles on the right. 
        }
        }
         \label{fig:QvalueEval}
\end{figure}

The advantage of using reinforcement learning for autonomous driving is that we can learn evaluation function for actions at any state. With classical planning, knowledge represented is not clear unless reading the code.  
Figure \ref{fig:QvalueEval} (left) shows a few sampled moments. 
The values for the Q values (outputs from the DQN networks) are printed in the caption.
Take the first moment for example, 
the ego vehicle was selecting the ``accelerate'' action because the action value corresponding to the acceleration action is the largest ($0.851$). So the acceleration action was chosen (according to the argmax operation over the Q values). 

Figure \ref{fig:QvalueEval} (right) shows the $Q(s, a=switch\_right)$ at a number of successive moments. 
The left color plots shows the values of switching right within the time window.
It clearly shows that the best moment of switching right is when the ego car moves near to the middle line between the two vehicles on the right. 
This finding means that our method has the potential to be used to learn and illustrate fine-grained driving knowledge that is conditioned on distances and speeds of other vehicles. 

% \begin{figure}[t]
%       \centering
%         \includegraphics[width=0.22\textwidth]{rsc/QAfinal2}
%         \caption{
% The left color plot shows the values of switching right within the time window:
% the middle moments have the largest values for switching right; while at the two ends, the values are small, indicating the switching right is not favorable because collision will occur. 
% The right color bar is the color legend. 
% The middle shows the trace of the car in the time window that corresponds to the left color plot (dotted line). 
% It shows that the best moment to switch right is near the middle line of the two vehicles on the right. 
% }
%         \label{fig:bestswitchingrightmoment}
%  \end{figure}

\section{Conclusion}\label{sec:conclusion}
In this paper, we studied an adversary driving scenario which is challenging in that the other vehicles may change lane to collide with our ego vehicle at a random time step.  
We proposed a novel way of combining classical planning methods with naturally defined primitive actions to form a set of single-step actionable options for reinforcement learning agents.
The key finding in this paper is that this method learns faster for collision avoidance and performs better than the primitive-action reinforcement learning agent. 
The comparison with human testers is promising, which shows our new method performs better than the average performance of three testers. 
A future work of this paper is to compare with human testers in a first-person view. 

%\bibliography{/home/hengshuai/reference}
\bibliographystyle{numbers}
\bibliographystyle{apalike}

\end{document}